# Highly dynamic physical interaction for robotics: design and control of an active remote center of compliance*

Christian Friedrich[1], Patrick Frank[1], Marco Santin[2], Matthias Haag[2]

***Abstract*— Robot interaction control is often limited to low dynamics or low flexibility, depending on whether an active or passive approach is chosen. In this work, we introduce a hybrid control scheme that combines the advantages of active and passive interaction control. To accomplish this, we propose the design of a novel Active Remote Center of Compliance (ARCC), which is based on a passive and active element which can be used to directly control the interaction forces. We introduce surrogate models for a dynamic comparison against purely robot-based interaction schemes. In a comparative validation, ARCC drastically improves the interaction dynamics, leading to an increase in the motion bandwidth of up to 31 times. We introduce further our control approach as well as the integration in the robot controller. Finally, we analyze ARCC on different industrial benchmarks like peg-in-hole, top-hat rail assembly and contour following problems and compare it against the state of the art, to highlight the dynamic and flexibility. The proposed system is especially suited if the application requires a low cycle time combined with a sensitive manipulation.**

## I. Introduction

A highly dynamic physical interaction is fundamental in many robotic applications. Particularly in contact-rich scenarios such as assembly, sensitive manipulation is required, which also entails a short cycle time. To allow on the one side sensitive-controlled manipulations and on the other a high dynamic, different control strategies exist, which are divided in passive and active interaction control [1]. Maybe the first ideas are based on passive compliant devices like the Remote Center of Compliance (RCC) [2], [3]. Bringing a compliant end-effector to a rigid robot is widely used in industry, because it allows a very cheap and efficient method for interaction control. The problem is that this control strategy struggles with inflexibility, because the RCC is designed to a predefined task. Beside the passive control strategies, the active interaction control deals with measuring the applied forces/torques on the robot end-effector and designing a control signal based on indirect control schemes like admittance, impedance control [4], [5] or direct ones like hybrid or parallel [6] force-torque-motion paradigms. All these approaches eliminate the drawback of inflexibility from the passive approaches, because the control parameters can be estimated automatically to the applied task [7], [8]. However, the main disadvantage of all these algorithms is that they have much lower dynamics, as the entire mass and inertia of the robot must be accelerated and decelerated when the robot encounters the environment.

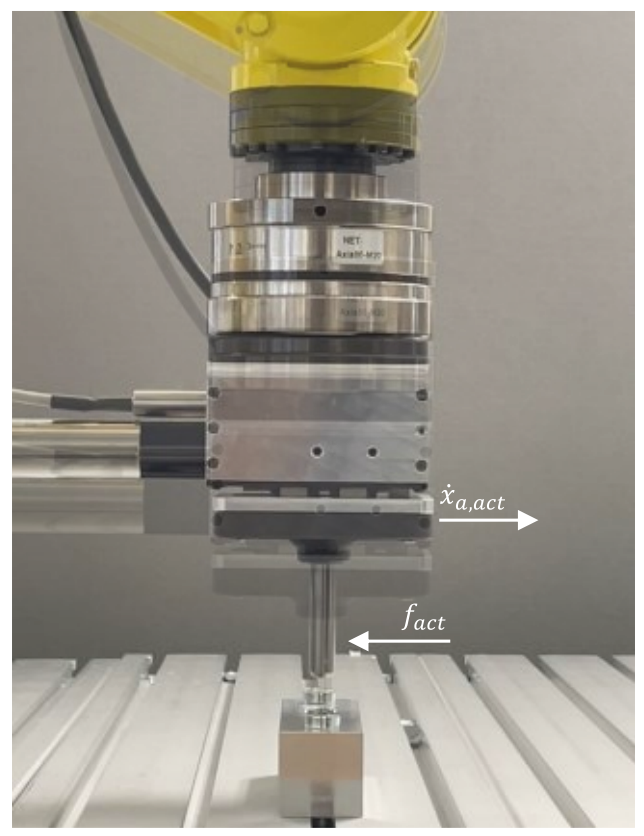

Figure 1. ARCC with active controlled axis during a PiH application.

A promising paradigm is the design of a compliant and actively controllable end-effector unit, which increases the bandwidth of the system due to its significantly lower mass and inertia than the entire robot structure. In this work we introduce a design and control paradigm for an Active Remote Center of Compliance (ARCC), which opens the door for high dynamic environment interaction with a high flexibility. Our key design is based on the well-known paradigm of a series elastic actuator [9], extended with new design and control concepts which consider the overall robotic system. This reduces the gap between active and passive interaction control, and we hope that this method can be widely applied in industrial robotic systems. We evaluate our proposed approach on different real-world applications and compare it directly to active control schemes.

## II. Related Work

Interaction control is a well-researched area within robotics. Most of the work is done in the separated fields of active and passive interaction control. The following section revisits related approaches from end-effector based interaction control (passive, semi-active) and robot-based interaction control (active), with a focus on assembly and disassembly tasks. We show the interconnections in the different domains and evaluate the advantages of bringing them together in a so-called hybrid interaction control approach.

### *A. End-effector based interaction control*

End-effector interaction control can be realized by passive and active systems. Passive systems are characterized by elastic elements that bend during a joining movement.

*This work was financed by the Baden-Württemberg Stiftung in the scope of the AUTONOMOUS ROBOTICS project ISASDeMoRo.

[1]Christian Friedrich and Patrick Frank are with the University of Applied Science Karlsruhe HKA, Moltkestraße 30, 76133 Karlsruhe, Germany (phone: +49 (0) 721 925-1723; e-mail: {christian.friedrich, patrick.frank}@h-ka.de). [2]Marco Santin and Matthias Haag are with the University of Applied Science Aalen, Beethovenstraße 1, 73430 Aalen, Germany (phone: +49 (0) 721 925-1723; e-mail: {marco.santin, matthias.haag}@hs-aalen.de).

Positional inaccuracies can thus be compensated. These passive systems are referred to as VRCC (variable remote center of compliance) [10]. In particular, systems are being developed to reduce stresses and deformations caused by positional inaccuracies. Many of them use elastomeric elements [11], [12]. Other approaches use flexure hinges to passively compensate inaccuracies at the gripper directly [13]. This compensation can also be achieved through various methods beyond the end-effector. Flexure hinges provide a versatile solution for numerous applications by enabling compliance across multiple degrees of freedom and accommodating complex bending [14], [15], [16].

The initial concept of an active RCC (ARCC) incorporates a force sensor, which enables the inference of forces exerted on the TCP [17]. In contrast, the design has no passive compliance of its own, which means that the bandwidth depends directly on the active drivetrain. Active systems use force control to maintain a constant or defined force. This idea has been taken up by a few current research projects and is applied particularly in the field of grinding. The decisive factor here is to grind a component contour with constant force [18], [19]. These systems only compensate for one degree of freedom. In combination with the force control, this ensures an effective grinding result. The drawback is that the position cannot directly controlled, which is crucial in sensitive scenarios.

### B. Robot-based interaction control

Most research that focuses on robot-based interaction control either uses impedance/admittance control, first introduced by Hogan [20], or hybrid position and force control, introduced by Raibert and Craig [21]. Especially in disassembly or assembly tasks, the focus is on impedance and/or admittance control due to their well-known dynamic properties and the possibility to deal with uncertainty in the selection of force- or position-controlled spaces.

Fan Zeng et al. [22] use an inner position and outer impedance control loop for peg-in-hole (PiH) tasks without the use of a force-torque-sensor (FTS). They identify and use the dynamic model of the robot to calculate the external forces and torques. Cho et al. [23] estimate the torque at each joint corresponding to external forces using a disturbance observer, which is then converted to the end-effector force using the robot Jacobian. This estimated force is used for impedance control in a PiH application. Mol et al. [24] achieve successful PiH insertions under large misalignments by using the KUKA LWRs impedance controller together with an outer admittance loop to adapt the impedance controllers setpoints. Stolt et al. [25] use force control to handle position to continuously switch and interpolate between impedance and admittance control. This combines the accuracy of admittance control in free space with the robustness properties of impedance control in stiff contact scenarios. Ott et al. [26] propose a hybrid system, which allows Yunlong Ma et al. [27] to combine visual servoing and admittance control to insert a serverboard into a server chassis. The work from Haddadin et al. [28] proposes a passivity-based framework for Cartesian force-impedance control which is also used in industry.

Our work focuses on the gap between robot and tool control to extend the interaction dynamic especially in industrial settings like assembly or disassembly. We introduce a novel design and control of our Active Remote Center of Compliance (ARCC). The main contributions of our work are:

- To the best of our knowledge, this is the first active remote center of compliance, which allows precise force and motion control in an embedded scheme.
- Introducing of a flexure hinge element for compliance, which can be easily scaled to different application requirements.
- Increasing the motion bandwidth up to 31 times against a purely robot-based controller.
- Establishing of surrogate models, which allow an objective comparison and offers the basis for an advanced controller design in future.

## III. Design and Control of the active RCC

The following section introduces the design of the ARCC. Further we compare the kinematic properties of our system against a robot-based control approach demonstrating the dynamic advantage. Based on the design we introduce the control scheme for the ARCC and the overall system with the robot.

### A. Design, modelling and kinematic properties

The design of ARCC adds an active and controllable component to the RCC, following a modular and compact approach. The system design parameters are defined based on a common industrial assembly process (e.g. small position uncertainties, medium interaction forces and low cycle times in a medium workspace). A positioning displacement of $\pm 2.5$ [mm] was selected, the assembly forces were set at a maximum of 60 [N], starting with lower forces in the tests and the cycle time is assumed to be 3 [$s$]. With a handling weight of 3 [kg], the ARCC should not increase the payload more than 0.6 [kg]. The system ensures that the TCP stays close to the robot flange, prohibiting larger interfering geometries and unfavorable disturbance torques on the robot. In addition, the system has a modular design due to its flatness and can therefore cover several degrees of freedom with additional modules. The modules are easily stackable and can be attached to common robot flanges and end-effectors using adapters.

The design features a high helix thread that is directly mounted to the brushless direct current servomotor (BLDC) via a rigid clutch (Fig. 2 (a), (b)). This solution offers a sufficiently low backlash for the application and is robust in the event of a collision. A controlled servomotor with the high helix thread enables precise positioning (theoretically $<$ 1 [µm]). The displacement measuring system is equipped with an LVDT (Linear Variable Differential Transformer) sensor, which enables very precise ($\leq \pm 30$ [µm]) non-contact measurement. The position measuring system detects the position of the follower module (passive part), e.g., the end effector, in relation to the position of the spindle nut (active part). This sensor is practically frictionless and ensures that the spring characteristic remains unaffected. Absolute measurements are also possible, and the long plunger travel prevents the displacement measuring system from being destroyed in the event of collisions.

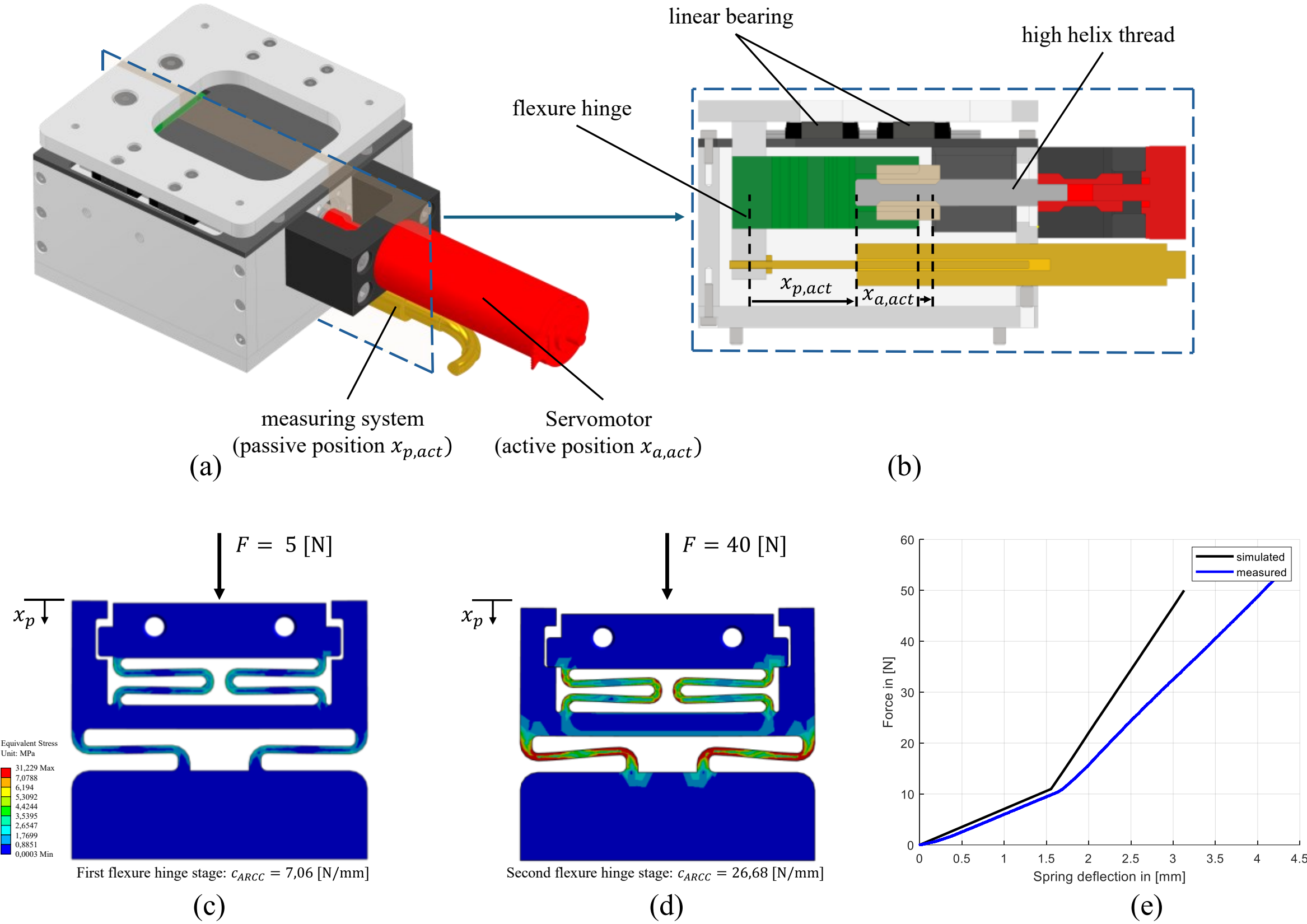

Figure 2. Design of the ARCC. Assembled design with the different components (a); section cut along the direction of motion (b); two-stage flexure hinge as compliant part of the ARCC (c), (d); position-force characteristic of the two-stage flexure hinge (e).

The flexibility of the system is achieved by flexure hinges. These have clear advantages over conventional springs. A very compact and flat design is possible and there are no disadvantages such as hysteresis and stick-slip effects due to the bearing of the springs. A flexure hinge can be attached directly to the threaded nut and the movable adapter plate. Only a linear bearing is currently required to block movements in unwanted directions. The flexure hinges can be easily scaled and replaced to handle different ranges of workpiece masses. A linear spring characteristic has proven to be very suitable but can cause problems in assembly cases and movements in which significantly different forces are used due to the short spring distance. The solution is based on a two-stage flexure hinge. This design works with a soft compliance in the initial area, which enables quick positioning and the detection of boundary areas. When a predefined deflection limit is reached, the hinge locks against a mechanical stop and enters a second spring stage. In this second phase, the hinge has a significantly higher spring rate, which enables a more rigid response. This is important for assembly or disassembly applications where a higher contact force is required. The principle is explained in Fig. 2 (c) and 2 (d) where different forces are applied, and the flexure hinge changes the spring characteristic. The different spring rates can be selected due to the requirements (workpiece load, robot, position tolerances, interaction task, etc.). Based on this, a parameterizable design model is used to create the flexure hinge, followed by additive manufacturing. Durability and possible changes in spring characteristics over time will be addressed in future work.

The transition point in the designed flexure hinge is at $\sim 11$ [N] with an elongation of $1.41$ [mm]. The linearity of the constants allows for their determination at arbitrary points within the respective stage. Based on Hook's law with $F = c \cdot x$, the spring characteristics for the used flexure hinge can be determined. For the selected applications we choose for the first stage $c_{ARCC,1} = 5\ [\mathrm{N}]/0.6439\ [\mathrm{mm}] \approx 7.06\ [\mathrm{N/mm}]$ and for the second stage $c_{ARCC,2} = 40\ [\mathrm{N}]/1.499\ [\mathrm{mm}] \approx 26.68\ [\mathrm{N/mm}]$. In Fig. 2 (e) the spring characteristic is shown, once simulated using FEM and once the real measurement. The deviation between the measurement and the simulation is based on the selected material parameters in the FEM software, the deviation from the filament used and the structure obtained in additive manufacturing. This is calibrated using a linear parameter.

Further, the kinematic properties of the ARCC are presented and compared against an industrial robot. Due to the selected components the ARCCs maximum travel is about 6 [mm] with velocities of up to 0.89 [m/s]. Theoretically, forces of up to 145 [N] can be applied by the drive unit. The model from the ARCC can be structured according to Fig. 3 (a), (b). To this, we describe the generated torque by the servomotor with $M_M$, induced by the desired motor velocity $\dot{x}_{a,des}$ which depends on the force controller and the desired force $f_{des}$. Further the helix thread is described by the inertia $J_s$, the gear ratio $i_s$ and the efficiency $\eta_s$. The kinematic coordinates from the active system are described with $x_a, \dot{x}_a, \ddot{x}_a$ and from the passive system with $x_p, \dot{x}_p, \ddot{x}_p$, as well as the corresponding masses with $m_a$ and $m_p$. The coupling between the active and passive system due to the flexure hinge is modelled with the spring rate $c_{ARCC}$ and a damping factor $d_{ARCC}$, which summarizes all dissipative effects in the system and the environment. The interaction with the environment is modelled with the spring rate $c_{env}$. The servomotor is controlled by a classic P-PI-PI cascade, whereby the magnitude optimum is used for the internal PI torque controller and the symmetrical optimum is applied for the PI velocity controller for parameterization purposes. This allows to model the transmission behavior from the desired motor velocity (active system) $\dot{x}_{a,des}(t)$ to the actual motor velocity $\dot{x}_{a,act}(t)$ with (1). Here $K_M$ is the motor gain and $T_M$ the motor time constant to be identified. This summarizes the electrical and mechanical motor properties and the high helix thread behavior with the spindle inertia $m_{ers,a} = J_s i_s^2 + m_a$.

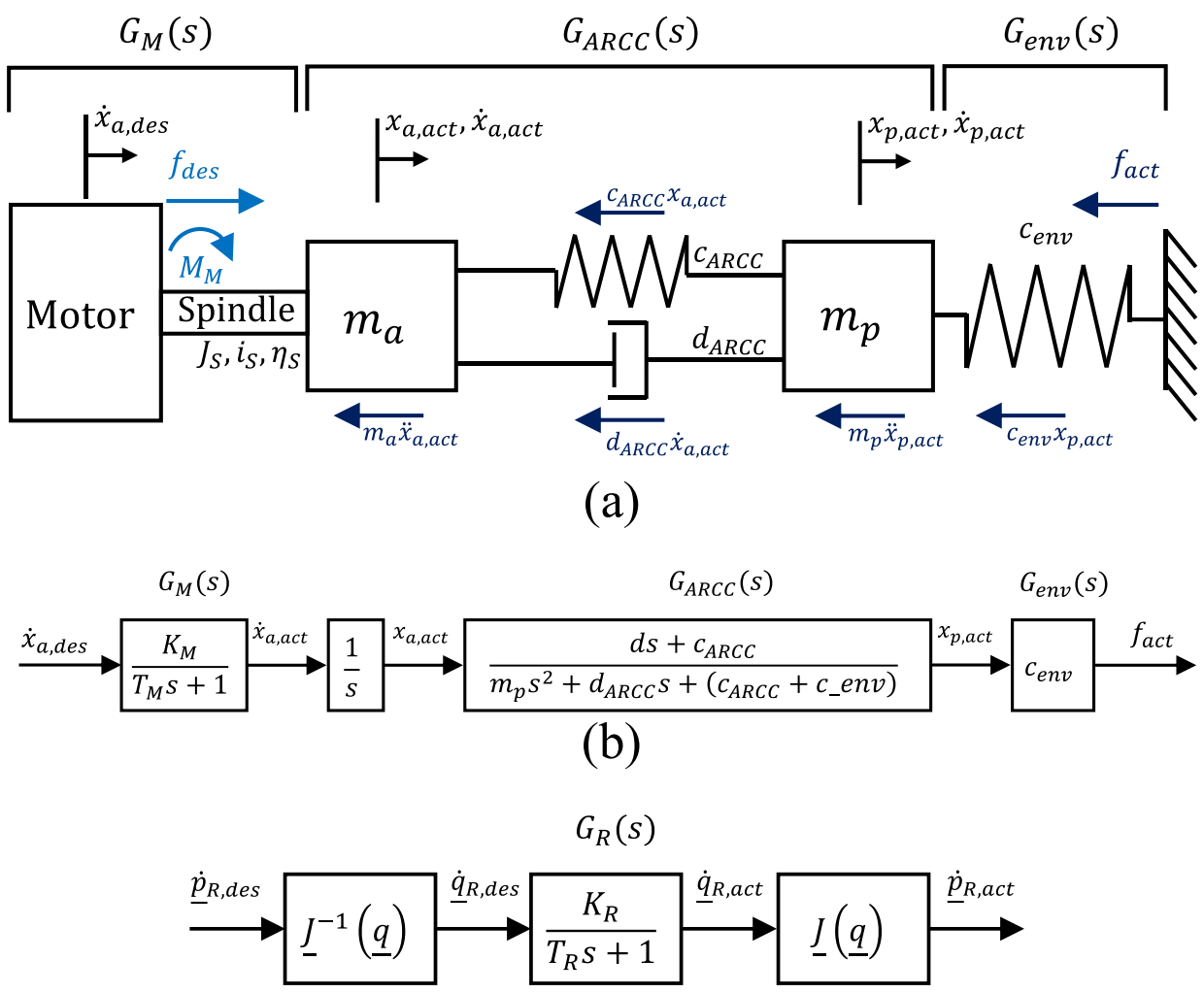

Figure 3. Model from the ARCC and the robot. Mechanic model ARCC (a); corresponding transfer function with servomotor $G_M(s)$, coupled masses $G_{ARCC}(s)$ and environment $G_{env}(s)$ (b); surrogate model robot (c).

$$G_M(s) = \frac{\dot{X}_{a,act}(s)}{\dot{X}_{a,des}(s)} = \frac{K_M}{T_M s + 1}. \tag{1}$$

The equations of motion from the ARCC, which describe the relationship between the current active position $x_{a,act}(t)$ and passive position $x_{p,act}(t)$ can be taken from Fig. 3 (a), resulting in (2) and the corresponding transfer function (3).

$$m_p\ddot{x}_p = -d_{ARCC}(\dot{x}_p - \dot{x}_a) - c_{ARCC}(x_p - x_a) - c_{env}x_p, \tag{2}$$

$$G_{ARCC}(s) = \frac{X_{p,act}(s)}{X_{a.act}(s)} = \frac{d_{ARCC}s + c_{ARCC}}{m_p s^2 + ds + (c_{ARCC} + c_{env})}. \tag{3}$$

Further, (4) is used to describe the interaction from the coupled position $x_{p,act}(t)$ from the ARCC with the environment and the resulting force $F_{act}(t)$.

$$G_{env}(s) = \frac{F_{act}(s)}{X_{p,act}(s)} = c_{env}. \tag{4}$$

For an objective comparison between ARCC and robot-based interaction control we introduce a simple Cartesian surrogate model for the robot, compare Fig. 3 (c). For a better overview we consider only one robot coordinate in the motion direction of the ARCC. Therefore, we assume that the desired Cartesian velocity from the robot $\dot{x}_{R,des}(t) \in \underline{\dot{p}}_{R,des}$ is transferred to the current robot velocity $\dot{x}_{R,act}(t) \in \underline{\dot{p}}_{R,act}$ via (5). In principial, the parameters $K_R, T_R$ from (5) depend on the robot configuration $\underline{q}$, because of the robot Jacobian $\underline{J}$. If the robot flexibility can be neglected (5) is valid. Further we consider only a sub workspace from the robot (nearly equal singular values $\sigma_i$, means $\sigma_i\left(\underline{q}_j\right) \approx \sigma_i\left(\underline{q}_{j+1}\right), \forall i,j$ from the Jacobian), which allows approximately to ignore this effect.

$$G_R(s) = \frac{\dot{X}_{R,act}\left(s,\underline{q}\right)}{\dot{X}_{R,des}\left(s,\underline{q}\right)} = \frac{K_R\left(\underline{q}\right)}{T_R\left(\underline{q}\right)s + 1} \approx \frac{K_R}{T_R s + 1} \tag{5}$$

From (5) it is straightforward to compare the kinematic properties between both approaches, because common industrial force control schemes are based on using the Cartesian velocity or position to control the interaction force. This underlying kinematic behavior significantly influences the overall dynamics in force control. For the identification of the subsystems (1), (3) and (5), a parametric approach is used, based on the instrument variable algorithm. As identification signal a sinus sweep is applied, with $f_{sweep,ARCC} = [10, \ldots, 120]$ [Hz] for the ARCC and $f_{sweep,R} = [1, \ldots, 10]$ [Hz] for the robot. Fig. 4 shows the identification results. The overall NRMSE in percent from the estimated models is 72% with a MSE from 0.274 [mm/s].

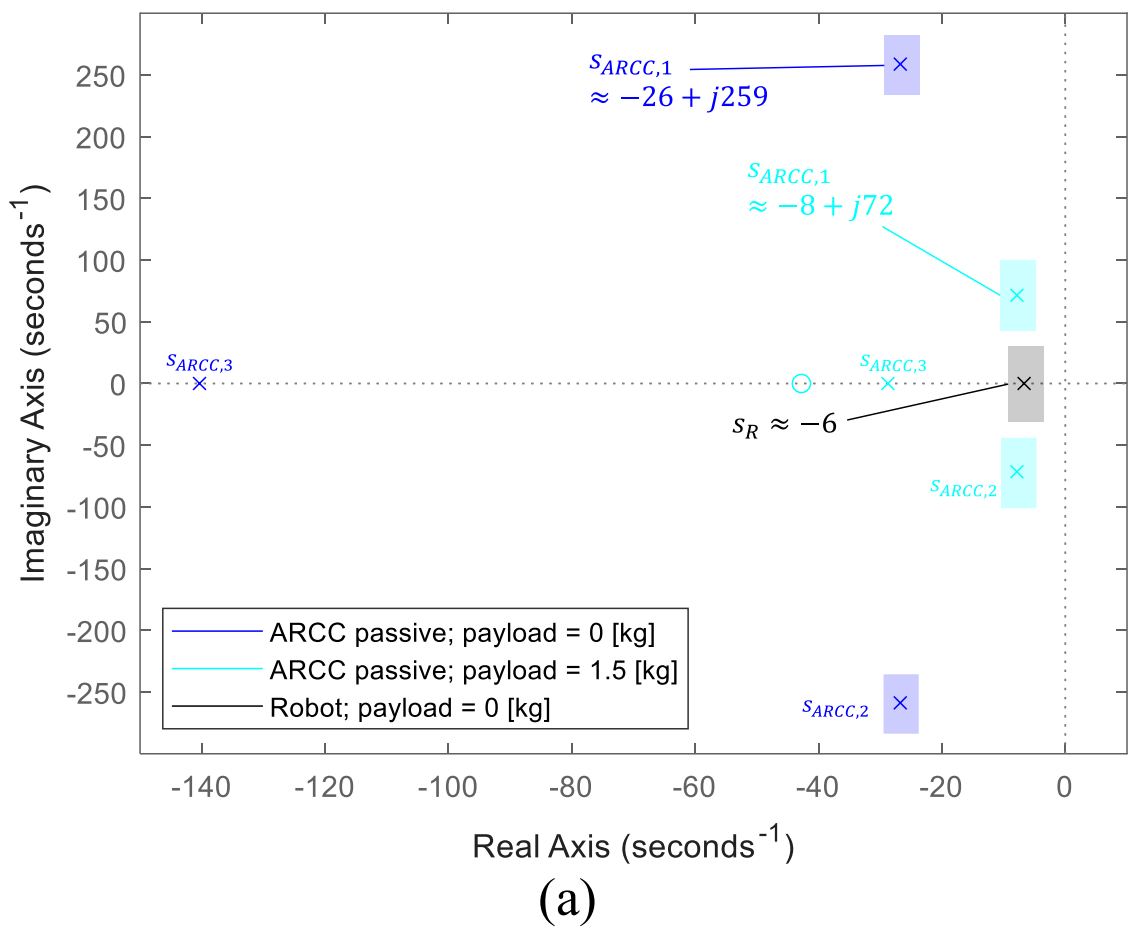

(a)

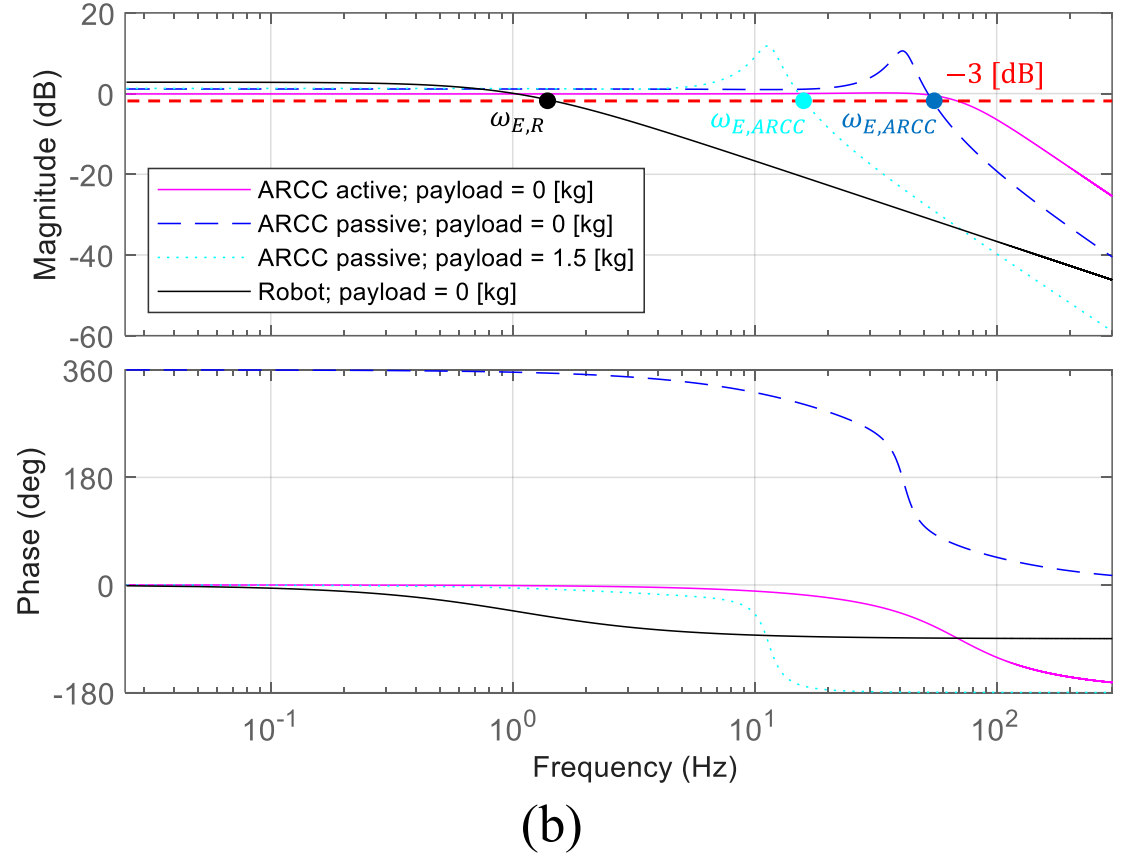

(b)

Figure 4. Results system identification using a single flexure hinge with $c_{ARCC} = 10$ [N/mm] and different payloads. Estimated poles and zeros for the parametric models from the active and passive part from the ARCC (3) and from the robot (5) (a); Bode diagram for the models (3) and (5) (b).

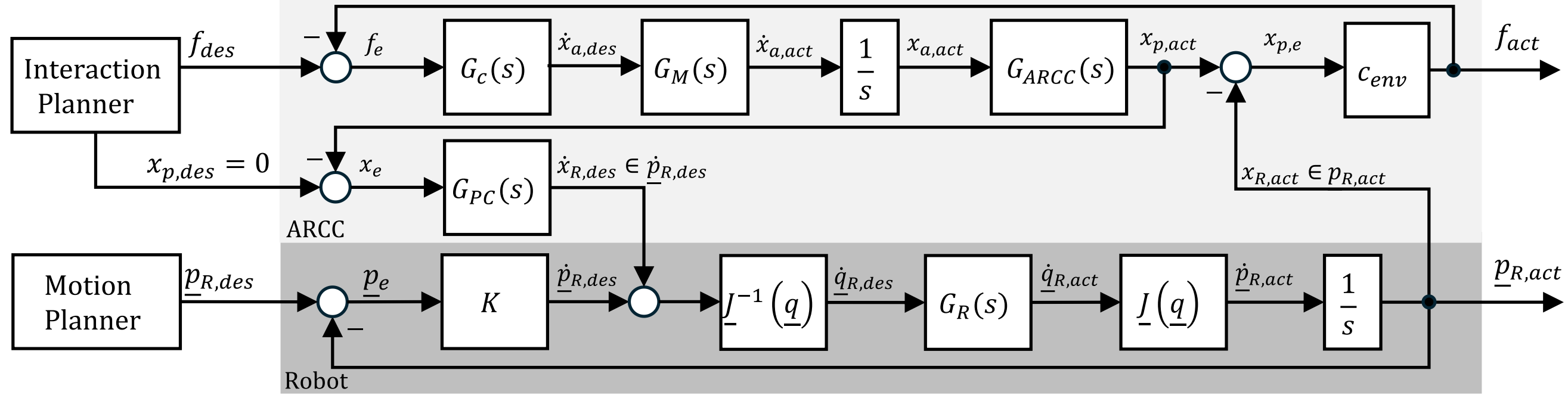

Figure 5. Overall control architecture ARCC with integration in the robot controller. The stiffness controller for the ARCC is extended with a parallel robot-based position controller to compensate position deviations in the direction of the ARCC.

For the estimation of the robot model (5), a sub workspace is selected with a homogenous and high transmission from the Jacobian. This sub workspace is also used for the conducted assembly experiments in IV. The robot is a FANUC LR Mate iD/7L, which fits the design parameters from III.A. The estimated values of the robot are not to be understood as absolute once. They only show the order of magnitude in the transmission behavior in relation to the ARCC. For robots in this class, the kinematic transmission characteristics are all in a similar range between a cut-off frequency $\omega_{E,R} \cong [1, \dots ,8]$ [Hz]. According to Fig. 4 (a) the pole from the robot (no payload robot) is between 1.3 times (1.5 [kg] payload ARCC) up to 4.3 times (no payload ARCC) slower than the poles from the ARCC (passive measuring system). The cut-off frequency from the robot is $\omega_{E,R} = 1.8$ [Hz] and from the ARCC between $\omega_{E,ARCC} = 16.6$ [Hz] (1.5 [kg] payload) and $\omega_{E,ARCC} = 55.9$ [Hz] (no payload). This significant increase the bandwidth from the force controller underlying motion system between 9 to 31 times against the robot with no additional payload.

### B. *Stiffness control with robot-based position compensation*

In this section we introduce the design of a simple stiffness controller for the ARCC in combination with a robot-based position compensation. The reason for using only a stiffness controller is that it allows a straightforward comparison with a robot-based interaction control approach, based on the introduced models in the previous section. The parameterization of the stiffness controller (6) for the robot and the ARCC is performed using a heuristic design via determination of the stability margin. The gain $c_c$ is a measure of the dynamics of the force controller, i.e. the smaller $c_c$ is, the faster the force controller reacts.

$$G_c(s) = \frac{\dot{X}_{a,des}(s)}{F_e(s)} = \frac{1}{c_c} = \tilde{c}_c \tag{6}$$

Because of uncertainties and process disturbance, the stiffness parameter is reduced by 10% from the stability margin. For the robot $\tilde{c}_{c,R} = 3.5 \cdot 10^{-4}$ [m/Ns] and for the ARCC $\tilde{c}_{c,ARCC} = 33 \cdot 10^{-4}$ [m/Ns] is determined, which is a result of the higher bandwidth of the ARCC. This further underlines the dynamic improvement. These parameters are applied in section IV. To allow a cooperation between the ARCC and the robot controller a second controller for the robot is introduced, compare Fig. 5. In addition to maintaining the target force, generated by the interaction planner, the second control objective is to ensure that the ARCC is in the center position, i.e. $x_{p,des}(t) = 0$. This ensures the functionality of the system even with large position deviations. The offset velocity $\dot{x}_{R,des}$ is generated by the controller for position compensation $G_{pc}(s)$, which results in a twist offset to the values from the robot motion planner.

## IV. Experimental Results

To evaluate our proposed ARCC in comparison to traditional force control using a stiffness controller (compare section III.B), we conduct three different experiments. The first one is a classic PiH experiment for fine manipulations. For the second experiment we mount a fuse on a top-hat rail to show the improvements in cycle time and for the last experiment we follow a wave contour with a desired contact force to show the dynamics. There are four configurations used in the following experiments. The first one is a traditional force control using the robot and an FTS. For the second one we extend the first one by the ARCC and disable the servo motor to mimic an RCC. The third and fourth configuration are using the ARCC with a one-stage flexure hinge and a two-stage flexure hinge respectively.

### A. *Peg-in-hole application*

Since our ARCC in its current version covers only one degree of freedom the hole for this experiment becomes a slot. All experiments were done using the traditional robot force controller in z-direction (downward direction). The x-direction (forward direction) is controlled using the respective configuration. For each configuration the PiH process is performed 20 times. As seen from the results in Tab. 1 the absolute force error for the ARCC is larger than the one from the robot configurations. The cause for this seems to be that the FTS also measures a force in x-direction when a torque is applied around the y-axis. The measurement of this additional force component is blocked by the ARCCs linear bearings. Future modules utilizing additional (rotational) degrees of freedom should account for this.

TABLE I. Results from PiH Experiment

| Configuration | Duration [s] | | Abs. force error [N] | |
|---|---|---|---|---|
| | **mean** | **std.dev.** | **mean** | **std.dev.** |
| Robot | 9.60 | 1.26 | 0.29 | 0.71 |
| Robot with RCC | 5.31 | 0.25 | 0.38 | 0.69 |
| ARCC (one-stage) | 5.84 | 1.26 | 0.82 | 0.45 |
| ARCC (two-stage) | 5.79 | 2.64 | 0.91 | 0.55 |

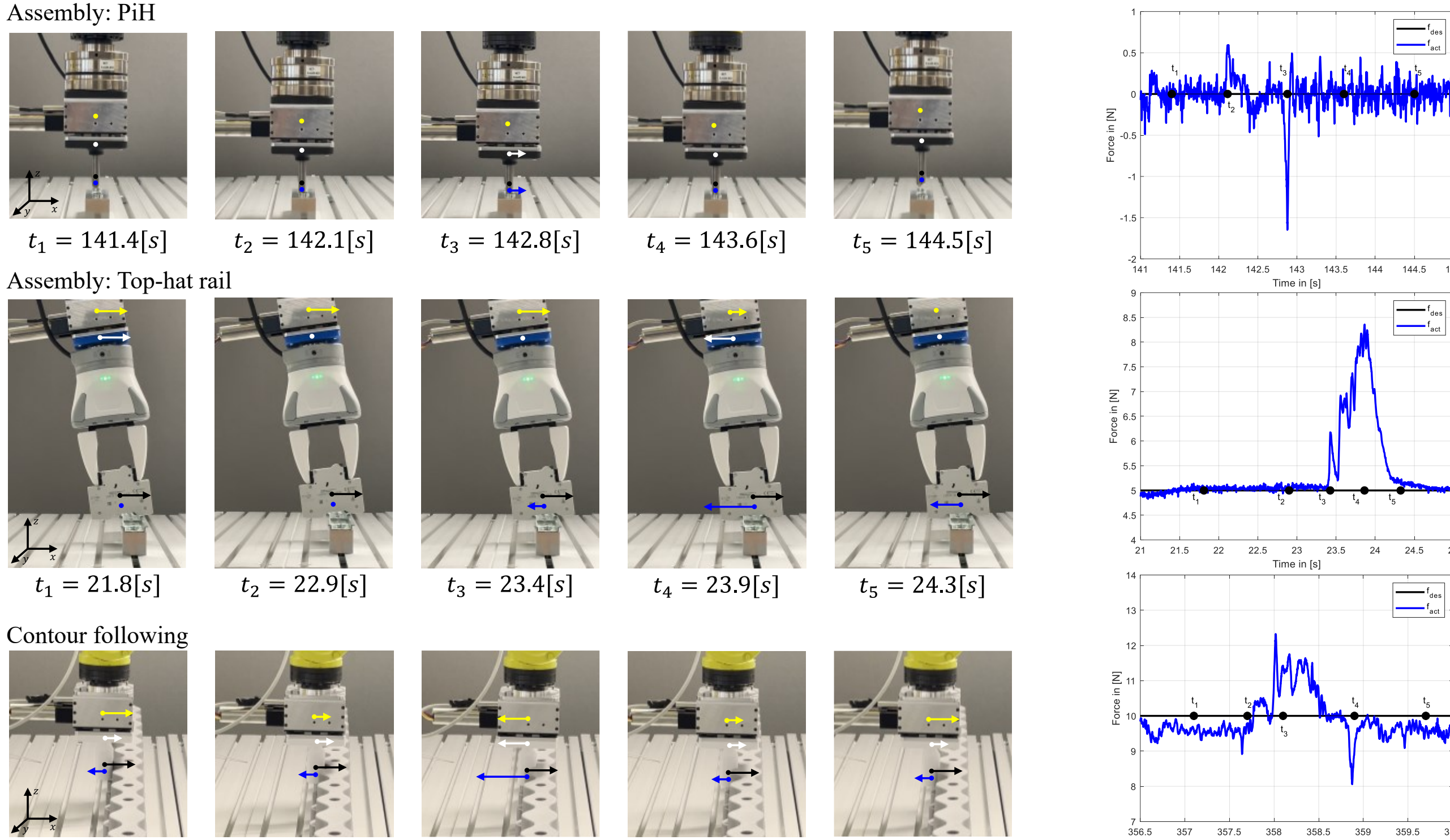

Figure 7. Validated applications and corresponding process force. Blue and black arrows indicate the magnitude of the actual (blue) and desired (black) force, respectively. White and yellow arrows indicate the velocity of the ARCC passive side and robot, respectively.

## B. Top-hat rail assembly

To mount a fuse onto a top-hat rail the robot first aligns the fuse roughly at a set offset ($x_{offset} = 15$ [mm]). After this the respective force controller is activated to reach contact with the rail ($F_{des} = 5$ [N]). The time from the first position to contact state is measured. This process is repeated 20 times for each configuration. Using the ARCC for this assembly step is around four times faster than the traditional approach while keeping the force overshoot reasonable, as shown by the results in Tab. 2. The force overshoot of our ARCC is greater than that of the traditional approach, which is a direct result from the parameterization according to III.B. If minimal force overshoot is the primary concern, $\tilde{c}_{c,ARCC}$ should be decreased at the expense of slightly lower dynamics.

TABLE II. RESULTS FROM TOP-HAT RAIL ASSEMBLY EXPERIMENT

| Configuration | Force [N] | Duration [s] | |
|---|---|---|---|
| | **max overshoot** | **mean** | **std.dev.** |
| Robot | 0.145 | 10.61 | 0.11 |
| Robot with RCC | 1.628 | 8.70 | 0.50 |
| ARCC (one-stage) | 1.869* | 2.68 | 0.18 |
| ARCC (two-stage) | 1.731* | 2.59 | 0.12 |

*The current version of the ARCC works by preloading the spring to the desired contact force, resulting in overshoots of 6.869 and 6.731 [N] respectively. For comparability we subtracted the preloading force from the results.

## C. Contour following

For this experiment we use 3D-printed contours with different amplitudes from 2.5 to 10 [mm] and a spacing of 65 [mm]. The desired contact force is $F_{des} = 10$ [N]. The robot speed along the contour is increased in 5 [mm/s] increments, starting at 5 [mm/s] until the contact between the end-effector (a 3D-printed pin) and the contour is lost. The results for the contours with an amplitude of 2.5 and 10 [mm] respectively are shown in Fig. 6 (a) and (b). The ARCC outperforms the traditional force control using a robot alone by a factor between 2.5 to 5 in terms of possible robot velocity while keeping the force error low. These results can be explained directly by the increased bandwidth of the underlying motion system compared to a robot with a much higher mass and inertia.

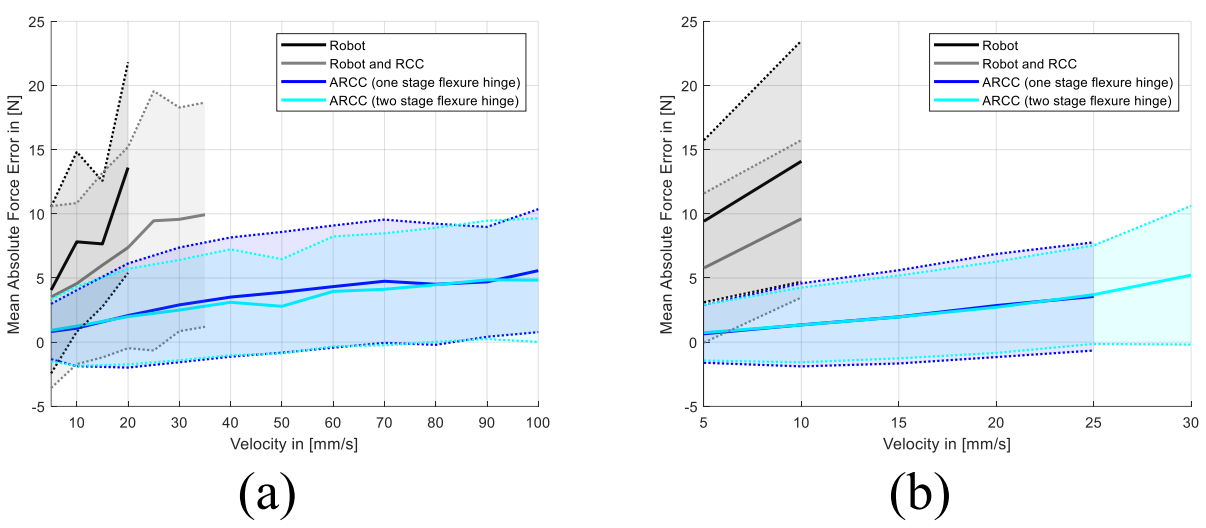

Figure 6. Results of the contour following experiment. Experiments were stopped when contact with the contour was lost. The lines show the mean absolute force error with the standard deviation indicated by the shaded area. (a) 2.5 [mm] amplitude wave contour. (b) 10 [mm] amplitude wave contour.

# V. DISCUSSION AND OUTLOOK

We introduce a novel active remote center of compliance and highlight the dynamic advantages in different experiments. The proposed system offers the same flexibility but with a higher dynamic compared to purely robot-based interaction control. This is achieved through the embedded controllable compliance which has a nearly algebraic behavior during contact force establishment. In future, we will expand our system to include further degrees of freedom and extend our modular approach while also reducing module size by optimizing the mechanical design. Also, we will design advanced control strategies to further increase the bandwidth of the complete system.